\documentclass[letterpaper]{article} 
\usepackage{aaai25}  
\usepackage{times}  
\usepackage{helvet}  
\usepackage{courier}  
\usepackage[hyphens]{url}  
\usepackage{graphicx} 
\usepackage{natbib}  
\usepackage{caption} 
\usepackage{algorithm}
\usepackage{algorithmic}

\usepackage{amsthm,amsmath,amssymb}
\usepackage{mathrsfs}

\usepackage{booktabs}

\usepackage{newfloat}
\usepackage{listings}
\DeclareCaptionStyle{ruled}{labelfont=normalfont,labelsep=colon,strut=off} 
\floatstyle{ruled}
\newfloat{listing}{tb}{lst}{}
\floatname{listing}{Listing}
\title{UN-DETR: Promoting Objectness Learning via Joint Supervision for Unknown Object Detection}
\author{
    Haomiao Liu\equalcontrib\textsuperscript{\rm 1},
    Hao Xu\equalcontrib\textsuperscript{\rm 1},
    Chuhuai Yue\equalcontrib\textsuperscript{\rm 1},
    Bo Ma\thanks{Corresponding author.}\textsuperscript{\rm 1},
}
\affiliations{
    \textsuperscript{\rm 1}Beijing Institute of Technology\\

    {ndwxhmz, xuhao\_cs, chuhuaiyue, bma000}@bit.edu.cn
}

\usepackage{bibentry}

\begin{document}

\maketitle

\begin{abstract}
Unknown Object Detection (UOD) aims to identify objects of unseen categories, differing from the traditional detection paradigm limited by the closed-world assumption. A key component of UOD is learning a generalized representation, i.e. objectness for both known and unknown categories to distinguish and localize objects from the background in a class-agnostic manner. However, previous methods obtain supervision signals for learning objectness in isolation from either localization or classification information, leading to poor performance for UOD. 
To address this issue, we propose a transformer-based UOD framework, UN-DETR. Based on this, we craft Instance Presence Score (IPS) to represent the probability of an object's presence. For the purpose of information complementarity, IPS employs a strategy of joint supervised learning, integrating attributes representing general objectness from the positional and the categorical latent space as supervision signals. To enhance IPS learning, we introduce a one-to-many assignment strategy to incorporate more supervision. Then, we propose Unbiased Query Selection to provide premium initial query vectors for the decoder. Additionally, we propose an IPS-guided post-process strategy to filter redundant boxes and correct classification predictions for known and unknown objects. Finally, we pretrain the entire UN-DETR in an unsupervised manner, in order to obtain objectness prior. Our UN-DETR is comprehensively evaluated on multiple UOD and known detection benchmarks, demonstrating its effectiveness and achieving state-of-the-art performance. Our method is available at https://github.com/ndwxhmzz/UN-DETR.

\end{abstract}

%

\section{Introduction}

Deep learning-based vision solutions have achieved remarkable success in the past \cite{krizhevsky2012imagenet, he2016deep, vaswani2017attention}, but their generalization performance in open scenarios still faces significant challenges. Within the closed-world assumption, conventional object detection frameworks \cite{ren2015faster, redmon2016you,carion2020end} are limited to detecting objects belonging to predefined categories present in the training set, thereby disregarding objects outside these predefined categories. This limitation leads to detrimental outcomes in real-world scenarios where high precision is essential. To advance from conventional detectors to open-world detectors, numerous outstanding works \cite{7780542, zheng2022towards} have emerged. Among them, Liang et al. (\citeyear{liang2023unknown}) clarified the definition of Unknown Object Detection (UOD). 

\begin{figure}[t]
\centering
\includegraphics[width=0.8\columnwidth]{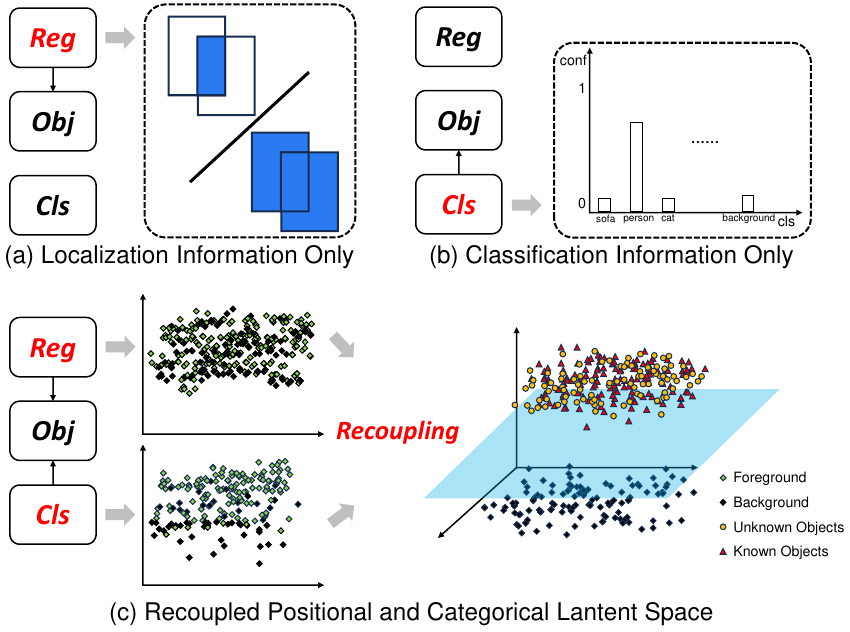} 
\caption{Joint supervision for objectness learning}
\label{fig1}
\end{figure}

UOD can be viewed as a two-step problem: 1) Locating all objects, both known and unknown, in a class-agnostic manner; and 2) Assigning specific categories to these objects. The first step hinges on how to learn the generalized features of any object, i.e., objectness, to distinguish them from the background. Difficulty arises because, under the definition of UOD, unknown objects lack supervision, and their objectness is mostly generalized from known categories. New methods \cite{wu2022two, liang2023unknown, zohar2023prob} have been developed to learn objectness, but they still struggle with low recall and precision, often misclassifying unknown objects for background or vice versa.

Our core insight is to explain the deficiencies of previous methods by their lack of considering both known category and location information in objectness learning. Classification information represents both class-specific categorization and the class-agnostic probability of being foreground. UnSniffer \cite{liang2023unknown}, only utilizes position prediction (IoU between predictions and ground truths)(Figure \ref{fig1}(a)), may misclassify high-IoU boxes without instances as foreground. Class-agnostic positional information is essential for accurate object localization. PROB \cite{zohar2023prob} ignores positional information and merges the objectness and classification head (Figure \ref{fig1}(b)), reducing localization accuracy. We hypothesize that extracting general objectness from complementary positional and categorical latent spaces may address the aforementioned issue.

To validate our theory, we developed UN-DETR, the first Transformer-based UOD framework, and introduced the Instance Presence Score (IPS), which integrates elements representing general objectness from positional and categorical latent spaces. Concretely, we introduce an IPS Predictor (IPP), alongside the original classification head and regression head, to directly output IPS. The optimization process is jointly supervised by signals from both spaces, ensuring their mutual complementarity. To enhance IPS learning, we propose a one-to-many assignment strategy that introduces more positive samples. Subsequently, we propose Unbiased Query Selection, to optimize the initialization of queries by replacing the original classification head with the learned IPP. Moreover, we propose an IPS-guided post-process strategy, filtering redundant boxes and further separating known from unknown objects. Finally, we pretrain the entire UN-DETR in an unsupervised manner using the region prior and the self-supervised encoder, to obtain objectness priors. Essentially, our approach improves the robustness and generalization of the detector, elevating UOD performance to a new level (Figure \ref{fig1}(c)).

To summarize, the contributions of our work are as follows:

\begin{itemize}
\item We reveal that a major flaw in previous UOD methods is the separate use of classification and localization information when learning objectness. To address this issue, we propose the very first Transformer-based UOD framework, UN-DETR. 

\item Our core design involves using a dedicated IPP to learn IPS under the joint supervision singals from complementary positional and categorical latent space. Moreover, IPS also participates in multiple stages of the UN-DETR, including query selection and post-processing.

\item Extensive experiments on both UOD and known detection benchmarks clearly demonstrate that our approach surpasses previous methods, achieving state-of-the-art performance.
\end{itemize}

\section{Related Work}

\subsection{Transformer-Based Detector}

Since Detection Transformer (DETR) \cite{carion2020end} pioneered the first fully end-to-end object detector, transformer-based detectors have gained significant attention for their outstanding performance and scalability. Deformable DETR (D-DETR) \cite{zhu2020deformable} further improved this by introducing deformable attention, which efficiently samples key elements, reducing computational demands, speeding up convergence, and enhancing performance.

To enhance training efficiency, recent methods \cite{li2022dn} have optimized the one-to-one assignment in DETR, which pairs each ground-truth object with a single prediction. Group-DETR \cite{chen2023group} implemented a group-wise one-to-many assignment, performing decoder self-attention within each group. Similarly, Co-DETR \cite{zong2023detrs} introduced a collaborative training scheme with multiple auxiliary heads using one-to-many label assignments.

Building on these advancements, we developed UN-DETR, the first transformer-based UOD method, based on D-DETR \cite{zhu2020deformable}. We apply one-to-many assignment specifically for IPS learning to facilitate generalized feature extraction from more positive sample queries. 

\subsection{Unknown Object Detection and Related Tasks}
Recent years have seen the emergence of tasks aimed at detecting unknown objects. Open Set Detection (OSD) \cite{7780542} requires identifying and excluding unknown samples, but issues with overconfidence affect accuracy. Open World Object Detection (OWOD) \cite{joseph2021towards} aims to detect both known and unknown objects, yet the absence of labels for unknowns prevents precise evaluation. Recently, Liang et al. (\citeyear{liang2023unknown}) further clarify the UOD evaluation protocol with both the precision and recall of unknown objects as metrics. 

Early OSD methods \cite{7780542,liang2018enhancing}, focus on distinguishing known and unknown objects. Techniques like maximum softmax probability \cite{hendrycks2017a}, minimum Mahalanobis distance \cite{denouden2018improving}, energy scores \cite{liu2020energy}, and virtual outliers \cite{du2022unknown,du2022vos} have been used. However, these methods primarily enhance known object detection, reducing unknown object recall. In contrast, Our approach seeks unbiased detection of unknowns.

Joseph et.al (\citeyear{joseph2021towards}) introduce OWOD with the ORE detector, featuring RPN-based unknown pseudo-labeling and contrast clustering. OW-DETR \cite{gupta2022ow} and others \cite{yang2022objects,zhao2023revisiting,wu2022uc,gupta2022ow,ma2023cat} explored various pseudo-labeling methods. Yet, pseudo-labeling often misclassifies non-objects as unknowns, reducing precision.

Recent efforts \cite{liang2023unknown, zohar2023prob, wu2022two} focus on objectness scores without pseudo-labeling, reducing false positives. \cite{wu2022two} extended ORE with a localization-based objectness head, improving recall. Similarly, \cite{liang2023unknown} introduced a localization-based GOC score using only known samples for supervision, and a graph-based boxes decision scheme. On the other hand, \cite{zohar2023prob} introduced a framework for classification-based objectness estimation that alternate between probability distribution estimation and objectness likelihood maximization of known objects in the embedded feature space, and ultimately estimating the objectness for different proposals.

In this paper, we propose a learnable objectness score, IPS, that integrates positional and categorical signals for robust objectness representation, avoiding pseudo-labeling pitfalls.

\begin{figure*}[t]
\centering
\includegraphics[width=0.8\textwidth]{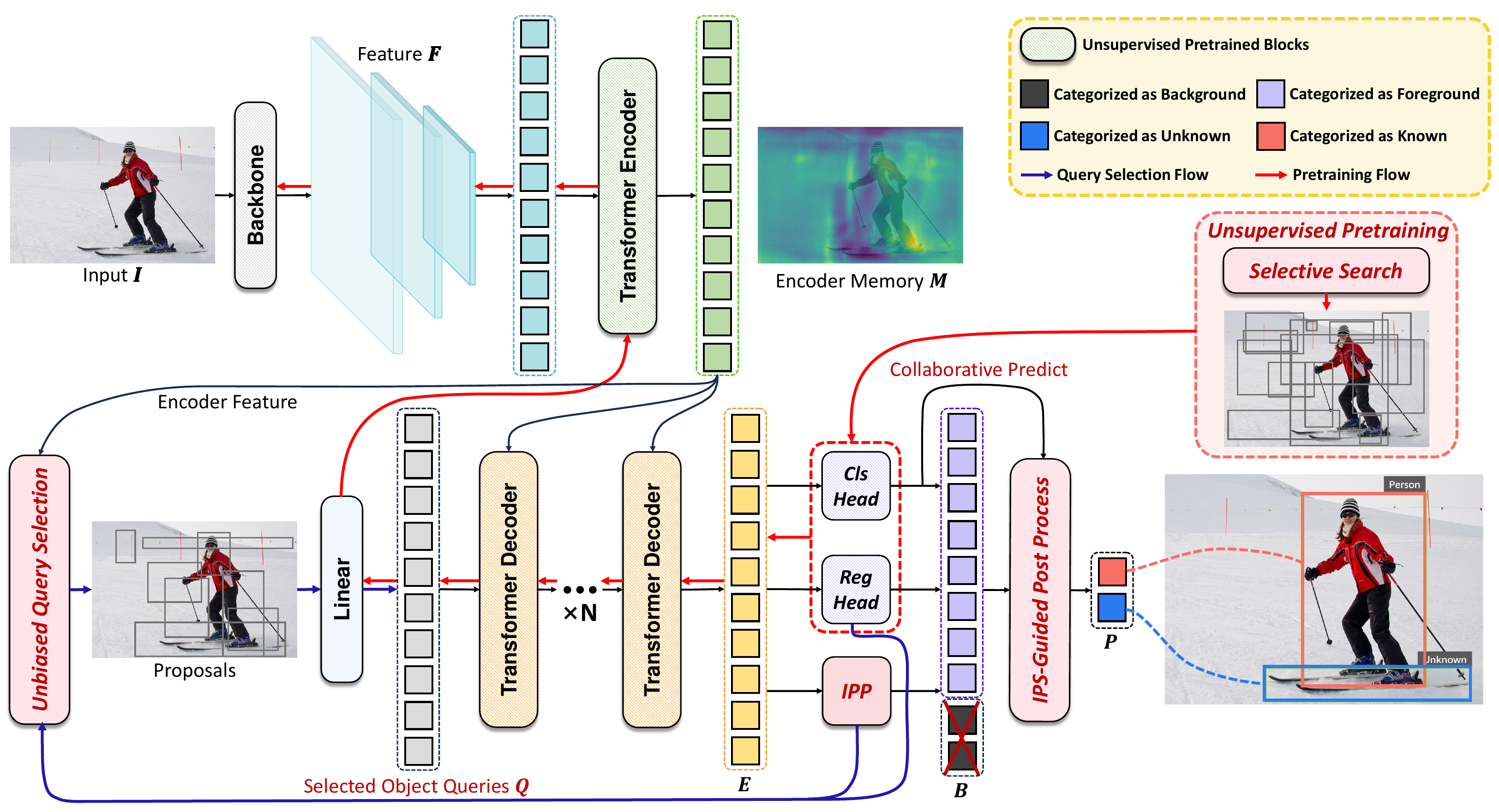} 
\caption{The overall architecture of UN-DETR}
\label{fig2}
\end{figure*}

\section{Preliminary}
\subsection{Two-Stage Deformable DETR Pipeline}
In D-DETR Pipeline, an input image $\boldsymbol{I}$ is processed by a backbone network to extract features, which are fed into an encoder using attention mechanisms to produce an enhanced feature sequence. In the decoder, $N_{\mathit{query}}$ object queries are updated through self-attention and cross-attention with the encoder's output, leading to refined queries $q{ \in} R^D$. These are then processed by the bounding box regression head $f_{\mathit{bbox}}$ and classification head $f_{\mathit{cls}}$ for final predictions. A one-to-one bipartite matching using $L_{\text{match}}$ ensures alignment with ground-truth (GT) labels for supervision. And two-stage D-DETR leverages region proposals generated by the encoder as initial object queries for further refinement in the decoder. The top-scoring region proposals are determined by applying $f_{\mathit{bbox}}$ and $f_{\mathit{cls}}$ to the encoder's output feature maps.

\subsection{Unknown Object Detection}
The task of unknown object detection represents an extension of conventional object detection frameworks. Referring to \cite{du2022vos,joseph2021towards}, the problem of unknown detection is formulated as follows. Given dataset $D=\{\boldsymbol{I},\boldsymbol{Y}\}$, where the $N$ input images are denoted as $\boldsymbol{I}=\{\mathbf{I}_1, \ldots, \mathbf{I}_N\}$, with corresponding labels $\boldsymbol{Y}=\{\mathbf{Y}_1, \ldots, \mathbf{Y}_N\}$. Each $\mathbf{Y}_i = \{y_1, \ldots, y_k\} (i{\in}\left[1,2,\dots{},N\right])$ contains a set of objects with $y_k = [l_k, b_k]$, where $l_k$ is the class label for bounding box $b_k$ represented by $x_k, y_k, w_k, h_k$. The known class set is denoted as $ \mathcal{K} = \{1, 2, \ldots, C\}$, and the unknown class is denoted as $\mathcal{U} = \{C + 1\}$. 

The model is trained on data labeled only with known-classes objects $\{(\mathbf{I}_n, \mathbf{Y}_n) $ $| l_k \in \mathcal{K}\}_{n=1}^{N_{\text{train}}}$, but tested on data including both known and unknown objects $\{(\mathbf{I}_n, \mathbf{Y}_n) | l_k \in$ $\mathcal{K} \cup \mathcal{U}\}_{n=1}^{N_{\text{test}}}$, where $N_{\text{train}}$ is the image number in the training set, $N_{\text{test}}$ for that in the test set, and $N = N_{\text{test}} + N_{\text{train}}$.

\section{UN-DETR}

\subsection{Overall Architecture}
The architecture of UN-DETR is depicted in Figure \ref{fig2}. The processing pipeline is as follows: An image $\boldsymbol{I}$ of dimensions $H \times W \times C$ is fed into the backbone to extract features $\boldsymbol{F}$. These features are then processed by the encoder to produce feature memory $\boldsymbol{M}$. The top K initial object queries are refined and filtered using $\boldsymbol{M}$, fed into the regression head, and a linear layer to produce $M$ object queries $\boldsymbol{Q}$. Subsequently, $N$ decoder layers transform $\boldsymbol{Q}$ into query embeddings $\boldsymbol{E}$, capturing the necessary spatial and semantic information for accurate Unknown Object Detection (UOD). These embeddings $\boldsymbol{E}$ are processed through three branches—classification, regression, and IPP—to predict potential instances. The predicted bounding boxes $\boldsymbol{B}$ undergo post-processing guided by IPS to yield the final prediction $\boldsymbol{P}$ of object instances. Prior to end-to-end training, the UN-DETR model undergoes unsupervised pretraining to establish objectness priors.

\subsection{Instance Presence Score Predictor}

\subsubsection{Instance Presence Score}

As we discussed in the Introduction, extracting representation from complementary positional and categorical latent spaces favors objectness learning. 
Disregarding class-agnostic categorical latent space, UnSniffer \cite{liang2023unknown} solely relies on position prediction (Intersection over Union, IoU, between the predictions and ground truth), may misclassify prediction boxes with high IoU scores but not containing any instances as foreground. Neglected by PROB \cite{zohar2023prob}, class-agnostic positional latent space directly impacts the detector's ability to locate potential objects. PROB only integrates the objectness head within the classification head but does not adjust the regression head, impacting the localization accuracy of unknown objects, leading to partial or oversized object prediction boxes.

\begin{figure}[t]
\centering
\includegraphics[width=0.8\columnwidth]{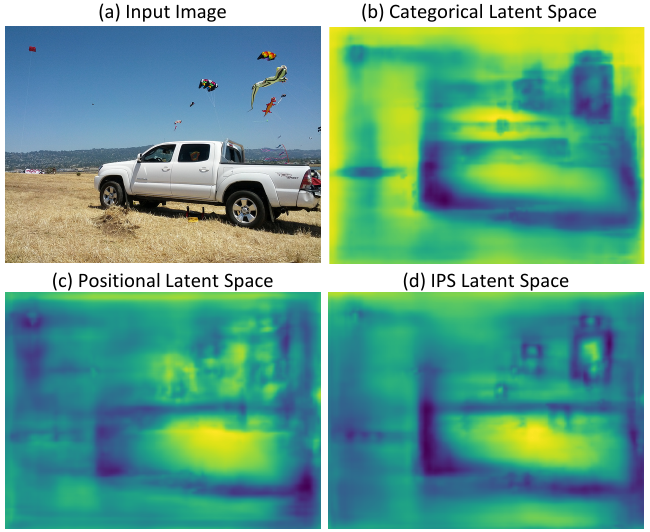} 
\caption{Visualizations of discriminability scores}
\label{fig3}
\end{figure}

Furthermore, to validate the above conjecture, we extract representations from either positional or categorical latent spaces as objectness score and visualize the discriminability score of feature maps during inference, as shown in Figure \ref{fig3}. Solely considering categorical latent space, as shown in Figure \ref{fig3}(b), the model exhibited higher discriminability score between instances and the background but poor distinction among instances themselves. Conversely, when using only the positional latent space representations, as illustrated in Figure \ref{fig3}(c), the model demonstrated greater distinctiveness among different instances but lesser between instances and the background. These experiments suggest that the two latent spaces are complementary in learning objectness.

In object detection task, predictions require locating objects (regression) and identifying them (classification). This highlights the need to integrate both positional and categorical latent spaces, suggesting that their combined use in the UOD task is more effective than treating them separately. Therefore, we formulate IPS by leveraging attributes of objectness from them, enhancing the use of knowledge learned from known categories and improving generalization to unknown objects. This approach also enhances the distinction between foreground and background, increasing robustness in diverse real-world environments, such as those with varying appearance and scale, which aligns with the primary goal of the UOD task.

Similarly, to validate the discriminability of IPS for different instances, the distinction maps are visualized as shown in Figure \ref{fig3}(d). After fully utilizing the representations from both latent spaces, instances are clearly distinguished from one another as well as from the background. The comparison clearly demonstrates effectiveness and superiority of IPS. 

\subsubsection{Jointly Supervised IPP}

To accurately estimate the IPS, we design a specialized IPP alongside the classification head and regression head. Specifically, the IPP is a simple single-layer feed-forward neural network and it inputs the query embedding $e_i \in \boldsymbol{E}$ and computes the corresponding IPS $I(e_i)$. 
For IPP training, we propose a jointly supervised strategy. First, the query embedding $e_i$ is fed into two heads $f_{\mathit{cls}}(e_i)$, $f_{\mathit{bbox}}(e_i)$ to obtain the representations $e_i$ in lower dimensions $e_{cls}, e_{bbox}$, which represent categorical and positional information obeying the two potential spaces $S_{cls}, S_{bbox}$, respectively. To remove the class-related components in $e_{cls}, e_{bbox}$, we extract the embedding composed of components representing generic objectness $e_{cls}^{o}, e_{bbox}^{o}$ from each of the two representations. For $e_{bbox}^{o}$, we compute generalized IoU (GIoU) after transforming it into a bounding box $\hat{b}_i$ with the matched GT $b_i$. GIoU is a metric based on spatial overlap that is independent of categories and thus performs better when confronted with unseen categories in the training set, and we formalize $e_{bbox}^{o}$ in terms of GIOU as follows: 
\begin{equation}
\hat{e}_{bbox, \sigma^{pos}}^{o} = \text{GIoU}(b_i, \hat{b}_{\sigma^{pos}})
\end{equation}
where $\sigma^{pos}$ is the index of the positive sample queries used to train IPP, as explained in the next subsection. 
For $e_{cls}^{o}$ , we represent the generalized objectness by the sum of $\mathcal{K}$ (number of known categories) logits, since the $\mathcal{K}$ logits represent the confidence of all the categories appearing in the training set, and the sum represents the overall probability of the foreground. This avoids the model's tendency to favor a particular category or categories, and reflects the robustness of the model to different categories of objects in various environments. So $e_{cls}^{o}$ is formalized as follows: 
\begin{equation}
\hat{e}_{cls, \sigma^{pos}}^{o} = \hat{P}_f{}_{\sigma^{pos}}
\end{equation}
where $P_f$ is the sum of $\mathcal{K}$ logits. 
To capitalize on the complementarity of $e_{bbox}^{o}, e_{cls}^{o}$, we set the objective probability $P_o(e_i) = \alpha \cdot e_{bbox}^{o} + \beta \cdot e_{cls}^{o}$, which serves as the supervised signal for IPP training.
Therefore, the loss function of IPP is as follows:
\begin{equation}
L_{\text{IPS}}^{\text{H}} = \ell_1(P_o, I(e_i)_{\sigma^{pos}}), 
\text{if } \text{GIoU}(b_i, \hat{b}_{\sigma^{pos}}) > \tau
\end{equation}
\begin{equation}
L_{\text{IPS}}^{\text{L}} = \ell_1(C, I(e_i)_{\sigma^{pos}}), 
\text{if } \text{GIoU}(b_i, \hat{b}_{\sigma^{pos}}) \leq \tau
\end{equation}
where $\ell_1$ denotes the L1 loss, $C$ is the objective constant and $\tau$ is the GIOU threshold. Here we introduce $C$ to increase the distinctiveness of IPS learning and maintain the stability of IPP training. The total IPS loss can be represented as:
\begin{equation}
L_{\text{IPS}} = L_{\text{IPS}}^{\text{H}} + L_{\text{IPS}}^{\text{L}}
\end{equation}

Finally, the entire loss of UN-DETR can be represented as:
\begin{equation}
L_{\text{UN-DETR}} = \lambda_1 \cdot L_{\text{IPS}} + \lambda_2 \cdot L_{\text{cls}} + \lambda_3 \cdot L_{\text{bbox}}
\end{equation}
where $L_{\text{cls}}$ and $L_{\text{bbox}}$ are consistent with the classification and regression loss of D-DETR, and $\lambda1$, $\lambda2$, and $\lambda3$ are the weights of the loss, set to 3, 2, and 5, respectively.


\subsubsection{One-to-Many Assignment}
In the Preliminary, we note that D-DETR employs one-to-one assignment to associate GT with potential objects. However, previous research has highlighted several issues with this approach, such as inefficient training. As a result, various one-to-many assignment methods have been proposed. These methods primarily aim to enhance convergence speed and training stability. Nonetheless, in UOD, beyond the aforementioned challenges, the most significant challenge is the invisibility of labels for unknown objects during training. Consequently, the joint supervision process of IPP must rely solely on features from the positional and categorical latent spaces. This reliance introduces potential uncertainty in the supervisory information, hindering the model parameters from iterating towards a more optimal solution, thereby compromising training stability and adversely affecting model performance.

One-to-many assignment allows for more flexible use of all supervision even when some of it is noisy or uncertain. By allowing multiple predictions to capture the same GT, the model aggregates information from these different matches, learning more stable and comprehensive object features.

Therefore, we propose a simple one-to-many assignment strategy to provide more positive samples for jointly supervised IPP training. Specifically, we introduce one set of sub-optimal queries besides one set of best-matching queries, since they also match known instances with high probability and are easily obtained by bilateral matching. The index for sub-optimal queries can be formalized as:
\begin{equation}
\hat{\sigma} = \arg \min_{\sigma \in \mathcal{S_N} / \hat{\sigma^*}}  \sum_{i} L_{\text{match}}(y_i, \hat{y}_{\sigma(i)})
\end{equation}
where $\hat{\sigma}$ is the index of the sub-optimal matching queries, which has the same length as the index $\hat{\sigma^*}$ of the best-matching queries. Then, the index of all positive sample queries for IPP training can be represented as:
\begin{equation}
\hat{\sigma^{pos}} = \hat{\sigma} \cup \hat{\sigma^*}
\end{equation}

Note that we only use the sub-optimal queries obtained from the one-to-many assignment for the jointly supervised IPP, and not for the classification and regression head. This is because the supervision for the regression head and classification head is derived directly from the labels and is inherently accurate. Introducing suboptimal supervision may therefore negatively impact their training. 

\subsection{Unbiased Query Selection}
To effectively select object queries relevant to the current input, the two-stage D-DETR employs an additional regression head along with a classification head to refine and filter appropriate proposals, initializing them as object queries. Specifically, this newly introduced classification head is trained using category labels represented as a tensor of all zeros, allowing the first dimension to indicate the probability that the input is a likely object. Consequently, during prediction, the top K proposals are selected based on the first dimension of the category prediction. This straightforward approach, however, disregards information from other dimensions predicted by the classification head, introducing a bias that causes this head to favor larger outputs in the first dimension. This bias leads to sparse gradient updates, as other classes contribute minimally, ultimately affecting the convergence and accuracy of this head, which is crucial to the overall detection task.

The additional classification head introduced in two-stage D-DETR essentially treats query selection as an instance recognition task, similar to the target of IPP. To address the bias mentioned earlier, we follow this approach and introduce an additional IPP, naming our method Unbiased Query Selection (UQS). In UQS, we replace the original classification head with an additional IPP to filter proposals. The supervision information of IPP reflects its class-agnostic nature, as its predictions denote the probability if the query represents a foreground object, independent of any category-specific priors. This lack of reliance on class-specific information eliminates potential bias, allowing IPP to focus solely on the objectness of the query itself, rather than being influenced by class-based assumptions.

\begin{figure}[t]
\centering
\includegraphics[width=0.8\columnwidth]{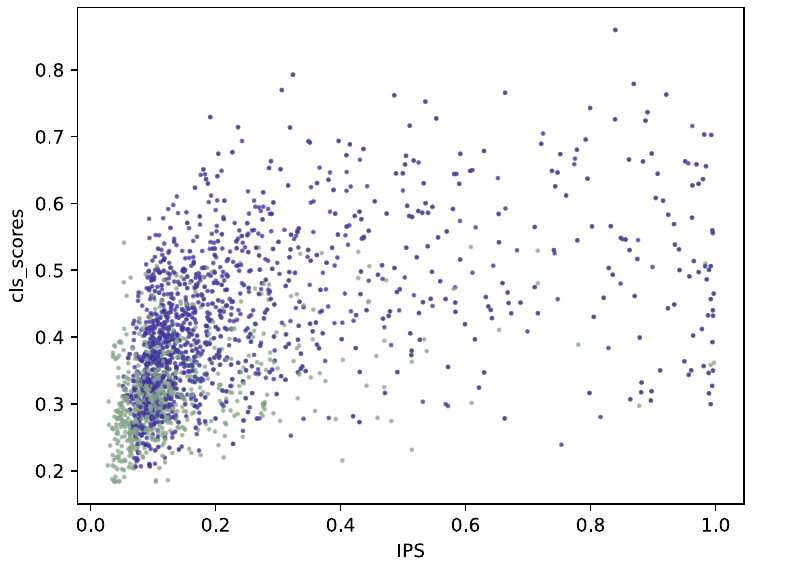} 
\caption{Visualization of classification scores and IPS for encoder features}
\label{fig4}
\end{figure}

To analyze the effectiveness of UQS, we visualize the classification scores and IPS of the encoder features in Figure \ref{fig4}. In the scatterplot, blue and green dots represent encoder features from models trained with Unbiased Query Selection and vanilla query selection, respectively. Dots closer to the top right indicate higher-quality features, meaning a greater likelihood of representing foreground objects. Notably, the top right corner has more blue dots than green, indicating that queries filtered by UQS are of higher quality and more likely to contain instances. Introducing the extra IPP, which incorporates positional information, allows for verifying the spatial accuracy of queries, reducing false detections. This mitigates the class-specific bias, improving the quality of object queries forwarded to the decoder. Additionally, integrating bias elimination into the loss function further optimizes gradients, enhancing training stability.

\subsection{IPS-Guided Post Process}

In D-DETR, the top-K bounding boxes are directly output as detection results. However, in UOD, unknown objects may outnumber known ones. A fixed K constrains the recall rate and is unreliable to set manually. Additionally, the original D-DETR post-processing doesn't work due to the one-to-many assignment in IPP training and the need to further differentiate unknown from known categories. To solve this, we propose the IPS-Guided Post Process, which includes IPS-Guided Non-Maximum Suppression (NMS) to remove redundant proposals and a dual-criteria unknown distinguish protocol.

\subsubsection{IPS-Guided NMS}

Traditional NMS methods\cite{neubeck2006efficient}  rank proposals based on classification confidence. However, when labels for unknown objects are missing, these methods may fail and even discard well-predicted boxes. Furthermore, \cite{jiang2018acquisition} highlight the inconsistency between localization and classification information in traditional NMS, where boxes with accurate localization may have low scores, or highly scored boxes may have poor localization. To eliminate as much of the background as possible and address the inconsistency, we propose IPS-Guided NMS. Considering the distance between center points of bounding boxes, we rank all boxes with IPS and calculate Distance IoU (DIoU) \cite{zheng2020distance} to measure overlap. 

\subsubsection{Dual-Criteria Unknown Distinguish Protocol}
Having identified all bounding boxes containing foreground objects, the next step is to classify the remaining proposals into known and unknown categories, after removing any redundant bounding boxes. Although there is no dedicated classification head for unknowns, utilizing IPS and classification confidence together still distinguishes between known and unknown. If both classification confidence and IPS are above set thresholds, the object is assigned to a known category. If classification confidence is low but IPS is high, the object is recognized but not confidently categorized, hence it's classified as unknown.

\subsection{Unsupervised Pretraining with Objectness Priors}
Self-supervised representation learning can reduce the amount of labeled data required by the model and improve its representation capability. We hope to improve the performance of UN-DETR by utilizing the related technology. To this end, following DETRreg \cite{bar2022detreg}, we pretrain the entire UN-DETR in an unsupervised manner to obtain objectness priors with both localization and classification. Specifically, we utilize an unsupervised region proposal generator, Selective Search \cite{uijlings2013selective}, to match object localization boxes. Moreover, we adopt a self-supervised image encoder, SwAV \cite{caron2020unsupervised}, to align the object embeddings used for classification. Note that to avoid introducing additional data, we only use the training set for unsupervised pretraining.

\section{Experiment}

Following the UOD Benchmark, we utilize COCO-OOD, COCO-Mixed \cite{liang2023unknown}, and VOC \cite{everingham2010pascal} as test sets and employ mAP, U-AP, U-F1, U-PRE, and U-REC as evaluation metrics, as detailed in the Appendix.\footnote{https://github.com/ndwxhmzz/UN-DETR.}

\subsection{Implementation Details}
In training, we use ResNet50 as the UN-DETR backbone. Moreover, the entire UN-DETR is pretrained on the VOC training set in an unsupervised manner \cite{bar2022detreg}. We introduce only one additional set of suboptimal queries for joint supervision of IPP training. The weight parameters $\alpha$ and $\beta$ are empirically set to 0.6 and 0.4, respectively. 
In Eq. 4, $C$ is set to 0.5 and $\tau$ is set to 0.6.

\begin{table}
    \centering
    \setlength{\tabcolsep}{1mm}
    \begin{tabular}{l|c|cccc}
        \toprule
        Methods & mAP & U-AP & U-F1 & U-PRE & U-REC \\
        \midrule
        MSP & 47.0 & 21.3 & 31.4 & 27.9 & 35.9 \\
        Mahalanobis & 44.7 & 12.9 & 27.1 & 30.9 & 24.1 \\
        Energy score & \textbf{47.4} & 21.3 & 30.8 & 26.0 & 37.7 \\
        VOS & 46.9 & 20.5 & 31.7 & 29.1 & 34.8 \\
        \midrule
        ORE & 24.3 & 21.4 & 25.5 & 16.3 & \textbf{78.2} \\
        OW-DETR & 42.0 & 3.3 & 5.6 & 3.0 & 38.0 \\
        \midrule
        PROB & 36.0 & 4.3 & 17.5 & 11.7 & 35.2 \\
        UnSniffer & 46.4 & \underline{45.4} & \underline{47.9} & \underline{43.3} & 53.5 \\
        \midrule
        \textbf{Ours} & \underline{47.2} & \textbf{47.0} & \textbf{54.9} & \textbf{54.5} & \underline{55.3} \\
        \bottomrule
    \end{tabular}
    \caption{Comparisons with other methods in the VOC-test and COCO-OOD datasets. The mAP is based on VOC-test, while the other metrics are from COCO-ODD. The best results are in \textbf{bold}, second best are \underline{underlined}.} 
    \label{table1}
\end{table}

\subsection{Results}

\textbf{Quantitative Analysis.} Tables \ref{table1} and \ref{table2} present the results of our method UN-DETR, alongside 8 classic or recent state-of-the-art methods, on the UOD Benchmark. Notably, on the COCO-OOD dataset, our UN-DETR outperforms others in metrics except for U-REC. Particularly for U-F1 and U-PRE, our method surpasses the second-best result by 7.0\% and 11.2\%, respectively. ORE's U-REC outperforms our method but also recall many non-objects. This is evident as our method's U-AP, U-F1, and U-PRE are all approximately twice as good as ORE's. On the COCO-Mixed dataset, our method maintains a lead in U-AP and U-PRE, exceeding the second-best result by 4.1\% and 0.5\%, respectively. The aforementioned results demonstrate that our UN-DETR surpasses the previously leading method in unknown object detection, attributable to our proposed jointly supervised IPP training. In addition, experimental results on both the VOC-test and COCO-Mixed datasets show that UN-DETR performs comparably to existing methods on known detections, which demonstrates that it does not improve unknown detections by sacrificing known detections. 

In general, our method outperforms other OSD methods as it is designed for detecting rather than excluding all unknown objects. Compared to pseudo-label-based OWOD detectors, our UN-DETR only introduces one additional set of query samples using one-to-many assignment strategy, which reduces the interference of negative samples and thus leads larger on U-PRE. Most importantly, benefiting from jointly supervised IPP training from both positional and categorical latent spaces, our approach exceeds other objectness-based approaches, as will be further demonstrated in the ablation study of Sec. 5.3.

\begin{table}
    \centering
    \setlength{\tabcolsep}{1mm}
    \begin{tabular}{l|c|cccc}
        \toprule
        Methods & mAP & U-AP & U-F1 & U-PRE & U-REC \\
        \midrule
        MSP & 36.4 & 5.5 & 16.9 & 19.0 & 15.3 \\
        Mahalanobis & 35.1 & 5.1 & 14.9 & 20.7 & 11.6 \\
        Energy score & 36.4 & 4.9 & 16.9 & 16.7 & 17.1 \\ 
        VOS & 36.4 & 5.1 & 17.2 & 18.4 & 16.3 \\
        \midrule
        ORE & 21.3 & 14.0 & 17.5 & 10.3 & \textbf{59.2} \\
        OW-DETR & \textbf{41.4} & 0.7 & 2.5 & 1.4 & 16.1 \\
        \midrule
        PROB  & \underline{40.1} & 9.4 & \underline{26.2} & 17.0 & \underline{56.7} \\
        UnSniffer* & 35.9 & \underline{14.8} & \textbf{26.7} & \underline{19.3} & 40.9 \\
        \midrule
        \textbf{Ours} & 34.0 & \textbf{18.9} & 24.7 & \textbf{19.8} & 32.8 \\
        \bottomrule
    \end{tabular}
    \caption{Comparisons with other methods in the COCO-Mixed datasets. * means that our replication results.}
    \label{table2}
\end{table}

\noindent \textbf{Qualitative Analysis.} Figure \ref{fig7} visualizes the results of various methods
. It is evident that our UN-DETR outperforms other methods both in localizing and identifying unknown objects. On the one hand, UnSniffer, which ignores category information, 
misidentifies some unknown objects, such as the elephant in the second row, and misses some obvious unknowns, such as the surfboard in the third row. 
On the other hand, PROB, which neglects positional information, has difficulty in accurately localizing the boundaries of objects, such as the water cup in the first row. On the contrary, our method accurately detects the most unknown objects, which benefits from the fact that we recouple generic objectness from both positional and categorical latent space. In addition, our method can accurately distinguish between known and unknown and exclude redundant boxes, which is attributed to our proposed IPS-guided post-processing. More visualization results are presented in the Appendix.

\begin{figure}[t]
\centering
\includegraphics[width=0.9\columnwidth]{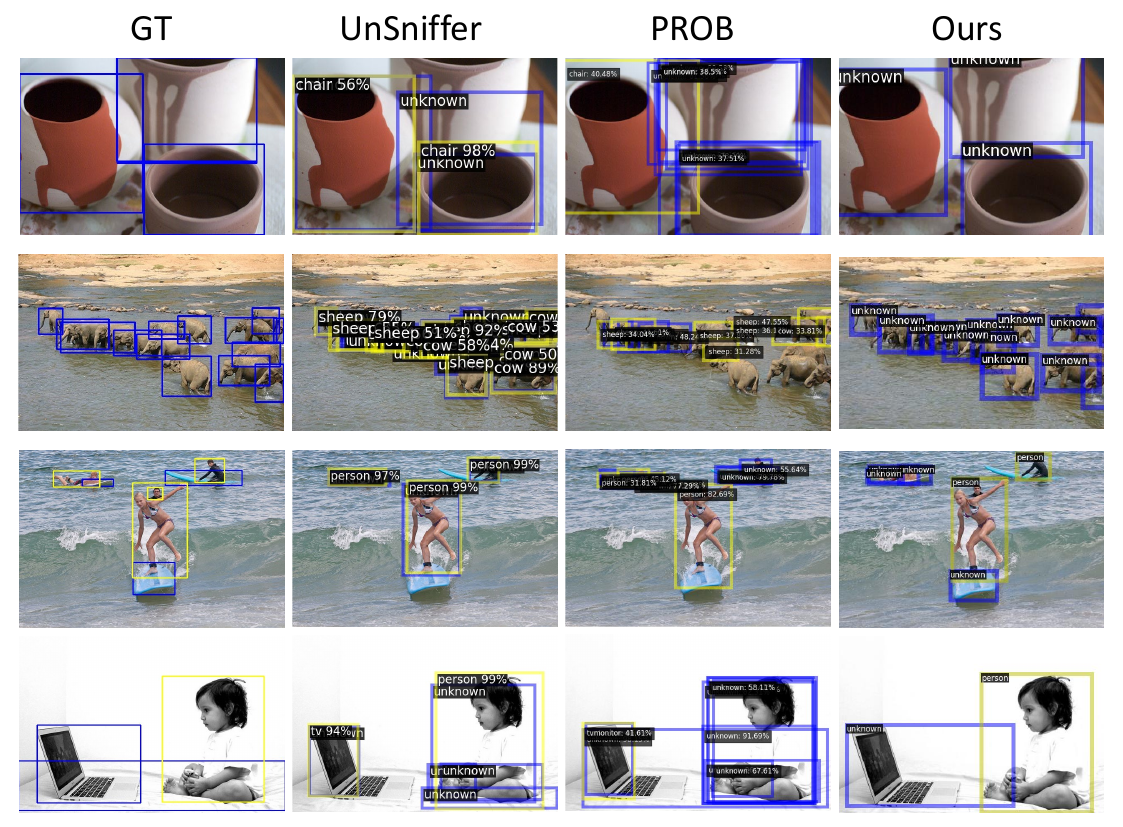}
\caption{Example results on COCO-OOD (first two rows) and COCO-Mixed (last two rows) datasets. Detections are overlaid on known (yellow) and unknown (blue) objects.}
\label{fig7}
\end{figure}

\subsection{Ablation Study}

To examine the contribution of each component in our method, we conduct adequate ablation experiments as presented in Table \ref{table3}. For IPP training, individual supervision from either the latent space of category or position significantly degrades the performance of the model (rows 1 and 2). If only the classification information is considered, the detector may miss some unknown objects and dramatically reduce the recall, while if only the regression information is focused on, the detector may confuse unknowns with knowns and recall non-objects, thus impairing the precision. Our UN-DETR trade-offs both of the above to obtain excellent precision and recall simultaneously.

In the one-to-many assignment strategy, we introduce only one additional set of samples, which outperforms the original one-to-one matching (row 3) and prevents the performance decrease from introducing more sets due to the possible negative sample noise (row 4).

For UQS, we train an additional IPP to replace the original classification head for query filtering, and the experimental results show that IPP outperforms not only the original classification head (row 5), but also the IPP supervised only with regression information (row 6). This proves the superiority of our joint supervision and the effectiveness of UQS.

To demonstrate the effectiveness of our proposed post-processing, we replace our IPS-Guided NMS with the original NMS (row 7), and the experimental results show that all the metrics of UN-DETR decrease. When unsupervised pretraining is not used (row 8), the U-PRE of UN-DETR decreases significantly. It's because pretraining provides a prior on objectness for the model, allowing it to initially acquire a certain class-agnostic perceptual ability after pre-training, which is crucial for UOD.

\begin{table}
    \centering
    \setlength{\tabcolsep}{1mm}
    \begin{tabular}{l|c|cccc}
        \toprule
        Row & Component & U-AP & U-F1 & U-PRE & U-REC \\
        \midrule
        1 & IPP only cls & 17.2 & 22.1 & \textbf{57.3} & 13.7 \\
        2 & IPP only reg & 40.9 & 23.9 & 14.7 & \textbf{63.4} \\
        \midrule
        3 & one-to-one & 46.0 & 51.3 & 51.4 & 51.1 \\
        4 & one-to-three & 46.0 & 53.7 & 53.5 & 53.9 \\
        \midrule
        5 & UQS origin cls & 39.7 & 51.1 & 56.7 & 46.4 \\
        6 & UQS only reg & 45.9 & 50.9 & 48.4 & 53.7 \\
        \midrule
        7 & w/o IPS-NMS & 46.7 & 52.3 & 50.1 & 56.8 \\
        \midrule
        8 & w/o Unsupervised & 45.8 & 44.5 & 35.0 & 60.1 \\
        \midrule
        9 & All & \textbf{47.0} & \textbf{54.9} & 54.5 & 55.3 \\
        \bottomrule
    \end{tabular}
    \caption{Ablation studies on COCO-OOD.}
    \label{table3}
\end{table}

\section{Conclusion}

We propose a novel transformer-based UOD method UN-DETR that outperforms existing state-of-the-art methods. We investigate the deficiencies of current methods in exploiting complementary classification and regression predictions, leading to unstable objectness learning. Therefore, the core insight of our approach is jointly supervised objectness IPS learning from both positional and categorical latent spaces. Then, we propose a one-to-many assignment strategy to provide more positive samples for IPS learning. Furthermore, IPS is employed for query selection and post-processing in UN-DETR due to its encoding class-agnostic categorization and localization information. Finally, we pretrain the entire UN-DETR in an unsupervised manner to obtain the objectness prior. We hope that our work will inspire further research in UOD within the community.

\section{Acknowledgments}
This work was supported by the Joint Funds of the National Natural Science Foundation of China (No. U2441206) and the National Natural Science Foundation of China (No. 62072042).
\bibliography{aaai25}

\clearpage\
\appendix
\section{Appendix}
This appendix is organized as follows: 1) More details of the method; 2) Benchmarks for UOD; 3) Experimental results, especially more visualizations; 4) Comparison with promptable approaches; 5) Limitations and future work.

\section{UN-DETR Details}

In this section, we show more details of the method.
In UQS:
\begin{equation}
I = f_{ipp}(Q)
\end{equation}
\begin{equation}
Q_{\text{UQS}} = \text{TopK}(I, K)
\end{equation}
where $Q$ denotes the output form Encoder, $K$ denotes the number of queries.

In one-to-many assignment:
\begin{equation}
Cost = \lambda_{1} \cdot Cost_{cls} + \lambda_{2} \cdot Cost_{box}
\end{equation}
\begin{equation}
Cost_{cls}(i, j) = -p_{cls}^{i}(c_j)
\end{equation}
\begin{equation}
Cost_{box}(i, j) = 1 - IoU(\mathbf{b}^{i}, \mathbf{b}_{gt}^{j})
\end{equation}
where $p_{cls}^{i}(c_j)$ denotes the probability that $i$-th prediction belongs to GT class $c_j$.

\section{Unknown Object Detection Benchmark}

\textbf{Datasets.} Following \cite{joseph2021towards,gupta2022ow}, we use the training set and validation set of the Pascal VOC datasets \cite{everingham2010pascal} as the training data that contains annotation of 20 known object categories. For testing, we use the three datasets as follows:

\begin{itemize}
  \setlength\itemsep{0pt}
  \setlength\parskip{0pt}
  \setlength\parsep{0pt}
  \item \textbf{PASCAL VOC datasets} \cite{everingham2010pascal} (the test set of it) are used to evaluate the accuracy of known object detection. 
  \item \textbf{COCO-OOD dataset} \cite{liang2023unknown} exclusively comprises unknown categories, comprising 504 images annotated with 1655 unknown objects. It serves for evaluating UOD.
  \item \textbf{COCO-Mixed dataset} \cite{liang2023unknown} comprises 897 images annotated with both known and unknown categories. It encompasses 2533 unknown objects and 2658 known objects. The inclusion of known objects in this dataset heightens the challenge of UOD. We evaluate the detection accuracy of known and unknown objects in this more challenging dataset.
\end{itemize}

\begin{figure}[t]
\centering
\includegraphics[width=\columnwidth]{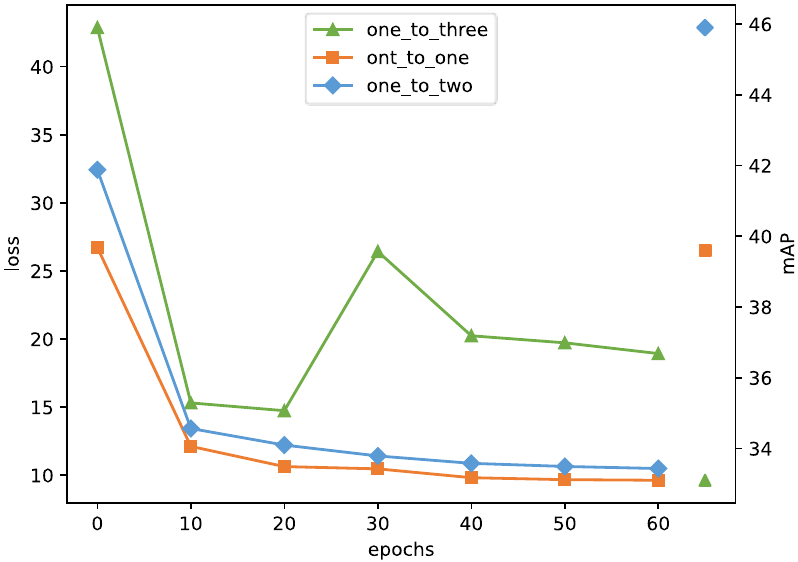} 
\caption{Convergence curves for different assignment strategies}
\label{fig5}
\end{figure}

\noindent \textbf{Evaluation Metrics.} To evaluate the performance of UOD, considering that \(TP_{u}\) denotes the true positive proposals of unknown classes, \(FN_{u}\) for false negative proposals, and \(FP_{u}\) for false positive proposals, we employ four unknown metrics as follows:

\begin{itemize}
  \setlength\itemsep{0pt}
  \setlength\parskip{0pt}
  \setlength\parsep{0pt}
  \item The definition of the Unknown Average Precision (U-AP) is consistent with the known object AP commonly used in conventional object detection \cite{everingham2010pascal}.
\item The Recall Rate of Unknown (U-REC) is defined as: 
\begin{align}
U\text{-}REC = \frac{TP_{u}}{TP_{u} + FN_{u}}
\end{align}
  
\item The Precision Rate of Unknown (U-PRE) is defined as:
\begin{align}
U\text{-}PRE = \frac{TP_{u}}{TP_{u} + FP_{u}}
\end{align}

\item For a comprehensive comparison, we report the Unknown F1-Score defined as the harmonic mean of U-PRE and U-REC:
 \begin{align}
U\text{-}F1 = \frac{2 \times U\text{-}PRE \times U\text{-}REC}{U\text{-}PRE + U\text{-}REC}
\end{align}
  
\end{itemize}

In addition, mAP is used to evaluate for known object detection. Note that we measure mAP over different IoU thresholds from 0.5 to 0.95. Other metrics are assessed at the IoU threshold of 0.5.

\section{Experiment Result}

\subsection{Ablation Study}
\textbf{One-to-many Assignment Strategy.} To demonstrate the improvement of the one-to-many assignment strategy for IPP training, we show the convergence curves for introducing different sets of query samples in Figure \ref{fig5}. It is clearly observed that compared to the one-to-one assignment, the introduction of one additional set of samples (one-to-two) significantly improves the model performance with a small difference in the convergence speed (see the right of Figure \ref{fig5}). In addition, the introduction of more sets of samples (one-to-three) significantly increases the instability of UN-DETR because of the possible presence of negative sample noise.

\noindent \textbf{Unsupervised Pre-training.} As shown in Table 3 of the paper, the unsupervised pre-training enhances the U-AP and U-PRE of our method. It's because it provides a prior on objectness for the model, allowing it to initially acquire a certain class-agnostic perceptual ability after pre-training, which is crucial for UOD. Notably, while others didn’t use our unsupervised pre-training, they used different pre-training methods. However, as indicated in the 8 th row of Table 3, our method still surpasses them in U-AP, even without pre-training. This demonstrates that our superior performance does not rely on pre-training. Futhermore, we retrained OW-DETR and PROB with the same pre-training method. As it is specifically suited for DETR-like architectures, we pre-trained Unsniffer using a closely similar approach. The U-AP of PROB has improved from 4.3 to 6.8, and the U-AP of OW-DETR has improved from 3.3 to 4.4. For Unsniffer, pre-trained weights achieved a U-AP of 26.1, showing strong unknown object perception, but final training yielded a U-AP of 43.9 due to method incompatibility.

\subsection{Qualitative Analysis}
\begin{figure}[t]
\centering
\includegraphics[width=\columnwidth]{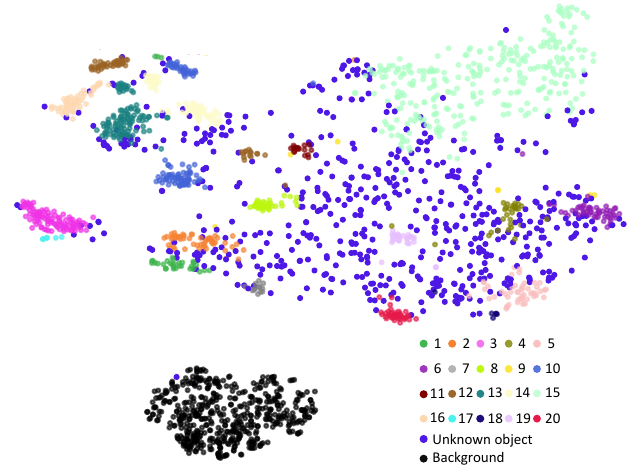} 
\caption{t-SNE visualization of various classes' hidden vectors}
\label{fig6}
\end{figure}

In order to verify the effectiveness of our approach, in particular the jointly supervised IPP, we visualized the output of the UN-DETR decoder using t-SNE. As shown in Figure \ref{fig6}, the unknown objects are scattered among the known ones, while the background is far from them. Thus, the t-SNE visualization intuitively demonstrates that our approach distinguishes between objects and non-objects, thanks to our proposed IPS that jointly re-couples generalized objectness from both positional and categorical latent spaces.

\begin{figure*}[t]
\centering
\includegraphics[width=\textwidth]{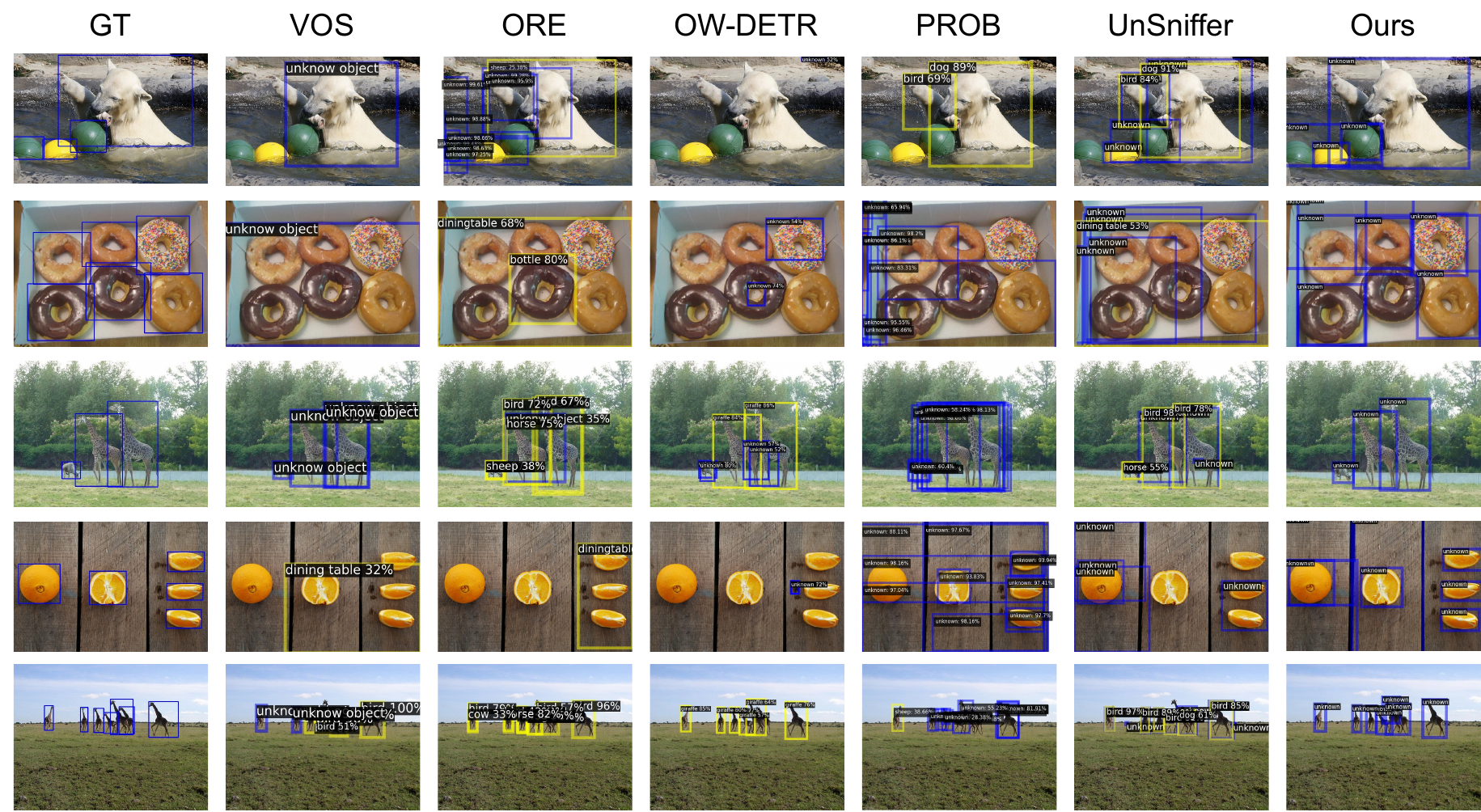} 
\caption{Example results on COCO-OOD datasets. Detections are overlaid on known (yellow) and unknown (blue) objects.}
\label{fig7}
\end{figure*}

\begin{figure*}[t]
\centering
\includegraphics[width=\textwidth]{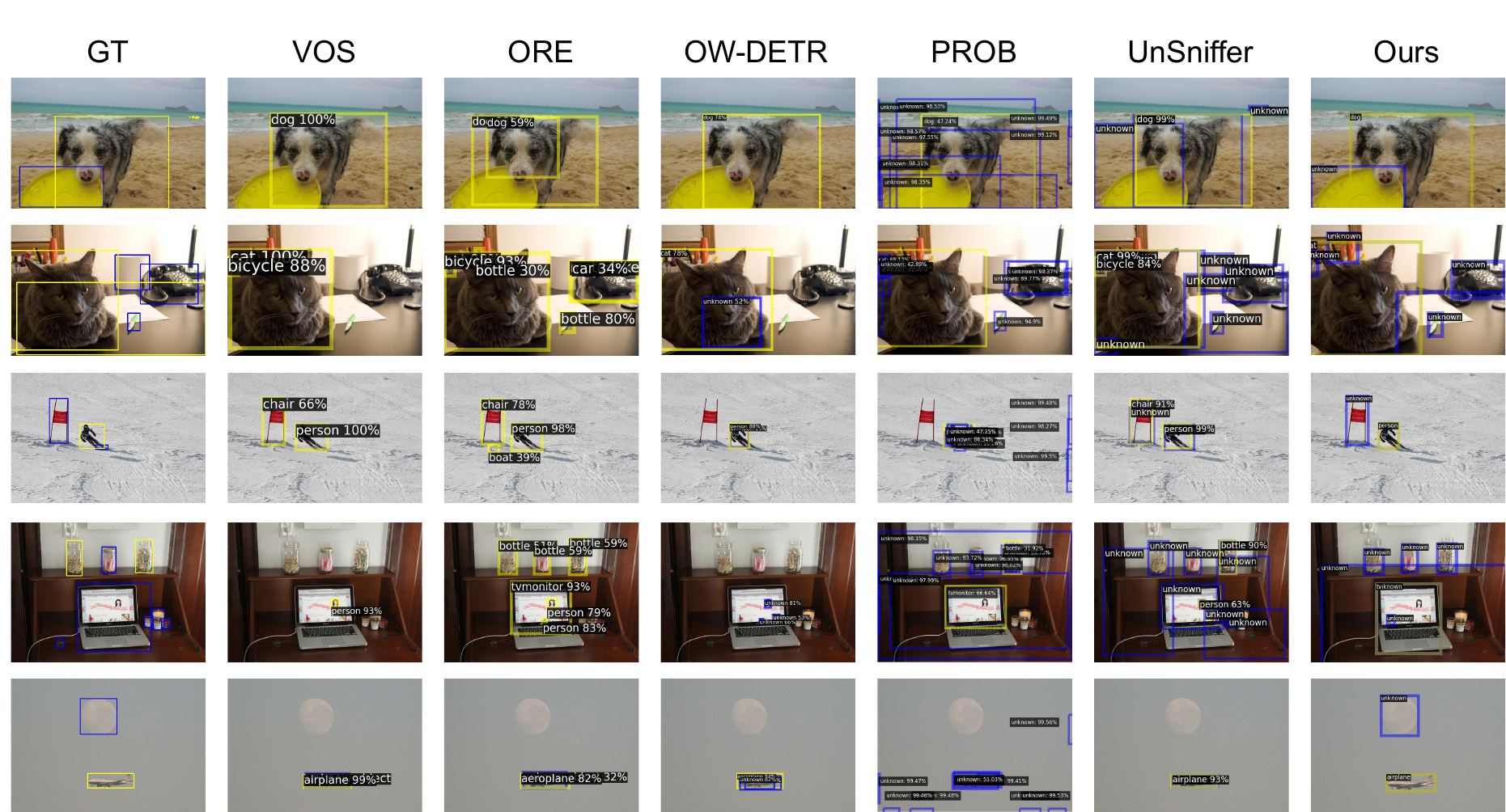} 
\caption{Example results on COCO-Mixed datasets. Detections are overlaid on known (yellow) and unknown (blue) objects.}
\label{fig8}
\end{figure*}

We show more visualization examples. Specifically, our UN-DETR is compared with 5 classic or recent state-of-the-art methods, including the OSD detector VOS \cite{du2022vos}, the pseudo-labeling-based OWOD detector ORE \cite{joseph2021towards} and OW-DETR \cite{gupta2022ow}, objectness-based methods PROB \cite{zohar2023prob} and UnSniffer \cite{liang2023unknown}. For a fair comparison, all results are either provided by the authors or reproduced by an open-source model trained on the same training set with the recommended setting.

Figure \ref{fig7} and Figure \ref{fig8} visualizes the results of above methods on example images from the COCO-OOD and COCO-Mixed datasets. Without learning reliable objectness, both OSD and OWOD methods misclassify or miss many apparently unknown objects. The objectness-based methods, on the other hand, despite detecting more unknown objects, are limited by individual supervision and thus still suffer from inaccurate object boundaries and low classification confidence. In contrast, our UN-DETR accurately detects the majority of unknown objects, thanks to jointly supervised IPP. For example, the donut in the second row and the orange in the fourth row of Figure 3, and the moon in the fifth row of Figure 4, these objects are only accurately predicted by our method. Moreover, attributed to our proposed IPS-guided post-processing, known and unknown objects can also be well distinguished, see the first and third rows of Figure 4.

\subsection{Computational Cost and Reasoning Efficiency}
Table \ref{table6} shows some results on computational cost and reasoning efficiency. Our method's computational cost and inference time are comparable to the original D-DETR and Faster-RCNN models, demonstrating its scalability and practicality. Additionally, testing on the CrowdHuman dataset, which averages 23 instances per image, yielded an FPS of 21.3. This shows our method's efficiency in large-scale, dense detection tasks.

\subsection{Result on OWOD Benchmark}
\cite{liang2023unknown} highlight COCO's incomplete labeling, limiting U-PRE assessment. PROB and OW-DETR, designed for OWOD \cite{joseph2021towards} focus on U-REC, leading to redundant boxes. Our method addresses both U-PRE and U-REC, achieving high U-AP and U-F1 on UOD benchmark\cite{liang2018enhancing}. Additionally, it scored 21.2 U-REC (MOWOD task1) and 18.8 U-REC (SOWOD task1) on OWOD, showing competitive performance.

\section{Comparison with promptable open-set detectors}
Recently, numerous promptable and grounded open-set detectors like RegionCLIP, GLIP and grounding DINO have been introduced, offering novel solutions for Object Detection in open scenarios. These are termed Visual-Language detectors because they utilize text prompts, unlike our Visual-Only detectors. While both can identify unknows, significant differences exist:

First, Learning Features. Prompt-based methods leverage multimodal data (image-text pairs) to align features, enabling generalization to unknowns, while our method leverages unimodal data to learn objectness features. Thus, direct comparison is not feasible.

Second, Data and Computational Requirements. Prompt-based methods (e.g., grounding-DINO) need large-scale, semantically-rich datasets (12M images and \textgreater 1000 categories) and significant computational resources (64 Nvidia A100 GPUs), whereas our method works with much smaller datasets (16K images, 20 categories) and a single RTX 4090.

Third, Category Limitations. Prompt-based methods require predefined text prompts, which may not cover all unknowns. Setting appropriate prompts for various scenarios is also challenging.

We prove our point later by experimental comparison with Grounding-DINO.


\begin{table}
    \centering
    \setlength{\tabcolsep}{1mm}
    \begin{tabular}{l|c|cccc}
        \toprule
        Methods & mAP & U-AP & U-F1 & U-PRE & U-REC \\
        \midrule
        Grounding DINO & \textbf{56.7} & 26.0 & 41.9 & \textbf{83.2} & 28.1 \\
        \textbf{Ours} & 47.2 & \textbf{47.0} & \textbf{54.9} & 54.5 & \textbf{55.3} \\
        \bottomrule
    \end{tabular}
    \caption{Comparisons with Grounding DINO in the VOC-test and COCO-OOD datasets.}
    \label{table4}
\end{table}

\begin{table}
    \centering
    \setlength{\tabcolsep}{1mm}
    \begin{tabular}{l|ccc}
        \toprule
        Methods & Parameters(M) & FLOPs(G) & FPS \\
        \midrule
        Faster-RCNN & 41.72 & 180 & 21.72 \\
        D-DETR & 42.02 & 137 & 24.33 \\
        \textbf{Ours} & 42.14 & 137.3 & 23.61 \\
        \bottomrule
    \end{tabular}
    \caption{Efficiency comparisons.}
    \label{table6}
\end{table}

\begin{table}
    \centering
    \setlength{\tabcolsep}{1mm}
    \begin{tabular}{l|cccc}
        \toprule
        Methods & U-AP & U-F1 & U-PRE & U-REC \\
        \midrule
        Grounding DINO & 26.0 & 41.9 & \textbf{83.2} & 28.1 \\
        Grounding DINO (3 classes) & 32.7 & 45.2 & 75.9 & 32.1 \\
        Grounding DINO (5 classes) & 41.7 & 52.7 & 75.8 & 40.4 \\
        Grounding DINO (7 classes) & 41.8 & 53.9 & 75.2 & 41.8 \\
        \textbf{Ours} & \textbf{47.0} & \textbf{54.9} & 54.5 & \textbf{55.3} \\
        \bottomrule
    \end{tabular}
    \caption{Comparison of Grounding DINO with different number of class prompts on COCO-OOD.}
    \label{table5}
\end{table}

\subsection{Experimental comparison with Grounding DINO}
Recently, \cite{2303.05499} designed a powerful “open set” object detector, Grounding DINO. To incorporate language information and enhance generalization to unseen objects, they introduced a tight modularity fusion technique based on DINO, enabling feature fusion at multiple levels of the pipeline. In addition, the author improved the training strategy of GLIP \cite{Li_2022_CVPR} by using a technique that utilizes sub level text features.

Liu et al. (2023) define the detection of arbitrary objects using manual inputs like category names or reference expressions as “open set object detection”. On the other hand, the UOD task requires the model to learn general objectness only from the known categories in the input image data and generalize to unknown objects, detecting instances that do not belong to any known category as “unknown”. Despite significant differences in task settings between our UN-DETR and Grounding DINO, the zero-shot inference ability of Grounding DINO is similar to UN-DETR's capability to detect unknown objects. Therefore, we evaluated Grounding DINO on the UOD benchmark for comparison with our model.

In our experiments, we use GroundingDINO-T with BERT-BASE as the text encoder. Referring to the prompt example given by Liu et al. (2023), we set the prompt to “Category 1. Category 2. ....... Category n. ”, and in particular, for the experiments on COCO-OOD, we set the prompt to ‘unknown object. ’.

Grounding DINO's zero-shot performance on the validation set of VOC achieves far superior results to other detectors, reflecting the splendid generalization performance resulting from the pre-training of large-scale datasets and the introduction of additional textual semantic information, as well as, the excellent performance of the DINO model structure itself.

Conversely, Grounding DINO's performance on COCO-OOD is significantly lower than that of UN-DETR, as shown in Table \ref{table4}, primarily because the task setting for UOD requires all objects to be uniformly labeled as "unknown". Grounding DINO's text encoder struggles to accurately encode the prompt “unknown. ” due to the inherent ambiguity of the concept, which contrasts with the more concrete and descriptive language typically used during its training. In contrast, UN-DETR is designed to correctly categorize known categories while classifying anything beyond that as "unknown". Grounding DINO, however, detects objects based on specific phrases or words provided in text prompts that indicate category information. This fundamental difference in task setting hinders Grounding DINO's performance in UOD tasks, leading to frequent missed detections. Nevertheless, it is worth noting that, owing to its powerful image feature extraction capabilities, Grounding DINO still outperforms some previous methods on COCO-OOD.

\begin{figure*}[!htbp]
\centering
\includegraphics[width=0.9\textwidth]{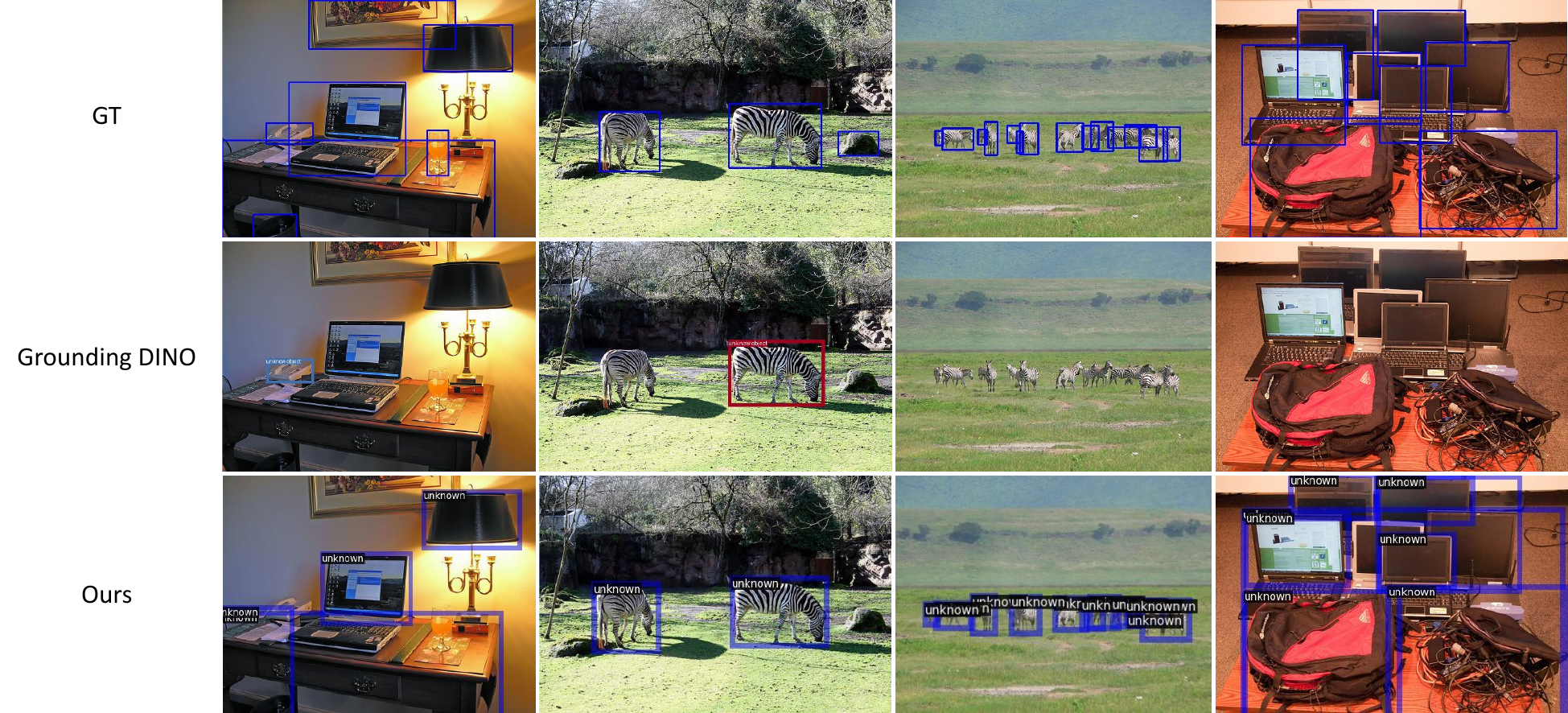} 
\caption{Example results on COCO-OOD datasets. Detections are overlaid on known (yellow) and unknown (blue) objects.}
\label{fig9}
\end{figure*}

Figure \ref{fig9} compares the visualization results of UN-DETR and Grounding-DINO on COCO-OOD. Due to limited category information from the text prompt, some obvious unknown objects are missed by Grounding-DINO, such as the lamp and table in the first column and the laptop in the fourth row, which directly reflects its limitations. On the other hand, our UN-DETR, which does not require any prompt input, detects all unknown objects.

To further assess Grounding DINO's limited performance in the UOD setting, we attempted to label several types of frequently occurring unknown objects in COCO-OOD within Grounding DINO's prompt, while fixing the model's output label as "unknown". The experimental results are shown in Table 2, adding category information in the prompt can improve its performance. This approach essentially predefines certain categories for Grounding DINO to detect, aiming to enhance its performance in this context. 

\begin{figure*}[!htbp]
\centering
\includegraphics[width=0.8\textwidth]{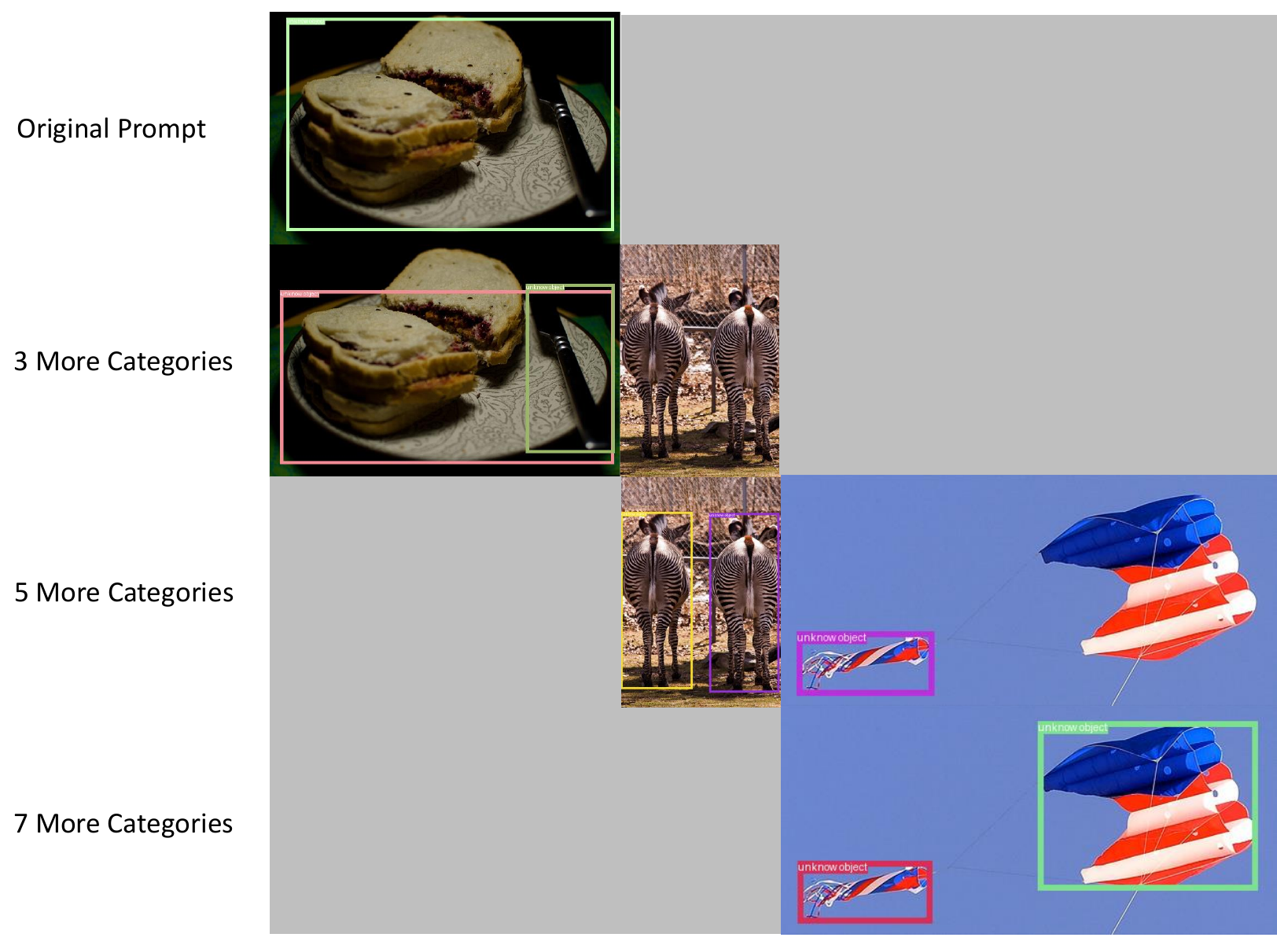} 
\caption{Example results with different number of class prompts on COCO-OOD datasets. Detections are overlaid on known (yellow) and unknown (blue) objects.}
\label{fig10}
\end{figure*}

As shown in Figure \ref{fig10}, as the category information contained in the prompt provided to Grounding DINO increases, the number of objects it can detect increases accordingly. This additional experiment demonstrates that Grounding DINO's performance is severely constrained by the information provided in the predefined prompt. If the prompt lacks an accurate and specific description, particularly in open-world scenarios where it is challenging to predefine all possible categories, Grounding DINO struggles to detect unknown targets accurately. This finding underscores both the research significance and the practical importance of the UOD task.

\section{Limitation and Future Work}
Although our method is fully evaluated on existing UOD benchmarks, its data size limits the method training and inference. In the future work, we will extend the unknown detection dataset for more comprehensive comparison of unknown detection. And, for application in real-world scenarios, we consider optimizing the inference speed of UN-DETR in future work.

\end{document}